\documentclass[runningheads]{llncs}

\usepackage{booktabs} % For formal tables

\usepackage[english]{babel}
\usepackage[utf8x]{inputenc}
\usepackage[T1]{fontenc}
\usepackage[]{multirow}
\usepackage{amsmath,algorithm}
\usepackage{algpseudocode}
\usepackage{graphicx}
\usepackage{changes}

\usepackage{subfig}
\usepackage[colorinlistoftodos]{todonotes}
\titlerunning{Optimisation and Illumination of WSRP Problems}

\begin{document}
\title{Optimisation and Illumination of a Real-world Workforce Scheduling and Routing Application  via Map-Elites}
\author{%
Neil Urquhart \and Emma Hart
}%

\institute{School of Computing, Edinburgh Napier University, Scotland, UK\\
\email{n.urquhart,e.hart@napier.ac.uk}
}

\maketitle

\begin{abstract}

Workforce Scheduling and Routing Problems (WSRP) are very common in many practical domains, and usually have a number of objectives. Illumination algorithms such as Map-Elites (ME) have recently gained traction in application to {\em design} problems, in providing multiple diverse solutions as well as illuminating  the solution space in terms of user-defined characteristics, but typically require significant computational effort to produce the solution archive. We investigate whether ME can provide an effective approach to solving WSRP, a {\em repetitive} problem in which solutions have to be produced quickly and often. The goals of the paper are two-fold. The first is to evaluate whether ME can provide solutions of competitive quality to an Evolutionary Algorithm (EA) in terms of a single objective function, and the second to examine its ability to provide a repertoire of solutions that maximise user choice. We find that very small computational budgets favour the EA in terms of quality, but ME outperforms the EA at larger budgets, provides a more diverse array of solutions, and lends insight to the end-user.

\end{abstract}

\section{Introduction}
\label{sec:intro}
Workforce scheduling and routing problems (WSRP) \cite{castillo-salazar-2012}  are challenging problems for organisations with staff working in areas including health care \cite{Braekers-2016} and  engineering \cite{Hart-2014}. Finding solutions is the responsibility of a planner within the organisation who will have an interest in the wider organisational policy decisions surrounding the solution. Such wider issues could include the implications of solutions with a lower environmental impact, the effects of switching to public transport, or the impact of changing the size of the workforce.

Multi-objective optimisation approaches are commonly used to find solutions, to WSRP instances, as they can provide a front of solutions that trade-off objectives \cite{Urquhart-2017}. However, fronts may only comprise a small section of the total solution space, and are difficult to visualise if there are many dimensions. Thus, it can be difficult for a planner to understand the range of solutions, why solutions were produced, and  in particular to know whether other compromise solutions might exist.

A class of algorithms known as {\em illumination} algorithms have recently been introduced  by Mouret {\em at al} \cite{Mouret-2015}, with a number of variants following, e.g. \cite{smith-2016,pugh2016}. Fundamentally different to a traditional search algorithm, the approach provides a holistic view of how high-performing solutions are distributed throughout a solution space \cite{Mouret-2015}. The method creates a map of high-performing solutions at each point in a space defined by dimensions of variation that are chosen by a user, according to characteristics of a solution that are of interest.  The resulting map (a Multi-dimensional Archive of Phenotypic Elites -) enables the user to gain specific insight into how combinations of characteristics of solutions correlate with performance, hence providing insight as well as multiple potential solutions. As the approach encourages diversity, it has often been shown to more capable of fully exploring a search-space, outperforming  state-of-the-art search algorithms given a single-objective, and can be particularly helpful in overcoming deception \cite{pugh2016searching}. We therefore hypothesise that an illumination algorithm might provide particular benefit to real-world problems such as WRSP, which contain multiple, and sometimes conflicting, objectives.  However, in contrast to the majority of previous applications of (ME) Map-Elites however which fall mainly in the domain of design problems (e.g. designing robot morphology), WSRP is a repetitive problem, which requires solving new instances repeatedly and obtaining acceptable solutions in reasonable time. While investing effort into producing an archive of solutions can pay off in a design domain, it may prove prohibitive for repetitive problems. Therefore, in the context of a WSRP based on the city of London, using real geographical locations and real transport information, we consider the following questions:

\begin{enumerate}
\item  How does the relative performance of ME compare to a standard Evolutionary Algorithm (EA) in terms of satisfying  a single objective-function over a range of evaluation budgets?
\item  Does MAP-Elites provide useful insights into problem characteristics from a real-world perspective through providing a range of diverse but hiqh-quality solutions?
\end{enumerate}

Using 10 realistic problem instances,  we demonstrate that for a small fixed evaluation budget, MAP-Elites does not outperform an EA in terms of the  objective function, but as the budget increases, it outperforms the EA on the majority of instances tested. Furthermore, even when it is outperformed by an EA in terms of the single objective, it can discover solutions that have better values for the individual characteristics. From a user-perspective, it may therefore present an acceptable trade-off between overall quality and insight.

\section{Previous work}

The Multi-dimensional Archive of Phenotypic Elites (MAP-Elites) was first introduced by Mouret {\em et al} \cite{Mouret-2015} and as discussed in the introduction, provides a mechanism for illuminating search spaces by creating an archive of high-performing solutions mapped onto solution characteristics defined by the user. To date, the majority of applications of illumination algorithms have been to {\em design} problems \cite{Mouret-2015,Vassiliades-2017}. Another tranche of work focuses on {\em behaviour} evolution in robotics, for example Cully {\em et al} \cite{cully2015robots}, who evolve a diverse set of behaviours for a single robot in a “pre-implementation” simulation phase: these are then used in future when the robot is in operation to guide intelligent choice of behaviour given changing environmental conditions.

To the best of our knowledge, an illumination algorithm has never been used to solve repetitive problems, i.e. problems faced in the real-world where acceptable solutions to problems have to be discovered in short time-frames, often many times a day. Typically these types of problems are combinatorial optimisation problems, e.g. scheduling, routing and packing, that often utilise indirect genotypic representations as a result of having to deal with multiple constraints. This contrasts to much of the existing work using MAP-Elites which uses a direct representation of design parameters (although the use of MAP-Elites with an indirect representation was discussed in \cite{Tarapore-2016}).

\section{Methodology}
\label{sec:methodology}

We consider a WSRP characterised by time-windows, multiple transport modes and service times, variations of this scenario  include the scheduling of health care workers as well as those providing other services such as environmental  health inspections.

We assume an organisation  has to service a set of clients, who each require a single visit. Each  of the visits $v$ must be allocated to  an employee, such that all clients are serviced, and an unlimited number of employees are available. Each visit $v$ is located at $g_v$, where $g$ represents a real UK post-code, has a service time $d_{v}$ and a time-window in which it must commence described by $\{e_{v}, l_{v}\}$, i.e. the earliest and latest time at which can start and finish.  Visits are grouped into {\em journeys}, where each  journey contains a subset $V_j$ of the $V$ visits and is allocated to an employee.  Each  journey $j$ starts and ends at the central office. Two modes of travel are available to employees: the first mode uses private transport (car), the second makes uses of available public-transport, encouraging more sustainable travel. The  overall goal is to minimise the total distance travelled across all journeys completed and forms the objective function for the problem. However, in addition, discussions with end-users \cite{Urquhart-2015} highlights four characteristics of solutions that are of interest:

\begin{itemize}
\item The total {\bf emissions} incurred by all employees over all visits
\item The total {\bf employee cost} the total cost (based on £/hour) of paying the workforce
\item The total {\bf travel cost} the cost of all of the travel activities undertaken by the workforce
\item The \% of employees using {\bf car travel}
\end{itemize}

We develop an algorithm based on Map-Elites to minimise the distance objective through projecting solutions onto a 4-dimensional map, with each axis representing one of the above characteristics.  Solution quality is compared to an Evolutionary Algorithm that uses exactly the same distance function as an objective, and an identical  representation, crossover and mutation operators.

Both the Map-Elites algorithm and the EA use an identical representation of the problem, previously described in \cite{Urquhart-2015}. The genotype defines a {\em grand-tour} \cite{Laporte-2013}, i.e. a single permutation of all $v$ required visits. This is subsequently divided into individual feasible journeys using a {\em decoder}. The genotype also includes $v$ additional genes that denote the model of transport to be used for the visit, i.e. public or private.

The decoder converts the single grand tour into a set of journeys to be undertaken by an employee. It examines each visit in the grand tour in order.  Initially, the first visit in the grand tour specified by the genotype is allocated to the first journey. The travel mode(car or public transport) associated with this visit in the genome is then allocated to the journey: this travel mode is then adopted for the entire journey (regardless of the information associated with a visit in the genome). The decoder then examines the next visit in the grand tour: this is added to the current journey if it is {\em feasible}. Feasibility requires that the employee arrives from the previous visit using the mode of transport allocated to the journey within the time window associated with the visit. Note that a travel mode cannot be switched during a journey. Subsequent visits are added using the journey mode until a hard constraint is violated, at which point the current journey is completed and a new journey initiated.

\subsection{The MAP-Elites Algorithm}
The implementation of MAP-Elites used in this paper is given in Algorithm \ref{alg:mapElites} and is taken directly from \cite{Mouret-2015}. The algorithm starts with an empty, N-dimensional map of elites: \{solutions $\mathcal{X}$ and their performances $\mathcal{P}$ \}. An initialisation method generates $G$ random-solutions; subsequent solutions are generated from elites stored in the archive. Following random selection of a solution (or solutions) from the archive, the {\em RandomVariation()} method applies either crossover followed by mutation, or just mutation, depending on the experiment.  All operators utilised are borrowed from \cite{Urquhart-2015}. The {\em mutation} operator moves a randomly selected entry in the grand-tour to another randomly selected point in the tour. The {\em crossover} operator selects a random section of the tour from parent-1 and copies it to the new solution. The missing elements in the child are copied from parent-2 in the order that they appear in parent-2. For each solution $x'$, a {\em feature-descriptor} $b$ is obtained by discretising the four features of interest associated with the solution (Section \ref{sec:methodology}) into 20 bins; for 4 dimensions this gives a map containing $20^4 = 160,000$ cells. The upper and lower bounds required for discretisation are taken as the  maximum and minimum values observed by \cite{Urquhart-2015} for each dimension during an extensive experimental investigation. A new solution is placed in the cell in the archive corresponding to $b$ if its fitness $p$ (calculated as total distance travelled) is better than the current solution stored, or the cell is currently empty.

\begin{algorithm}
\begin{algorithmic}
\Procedure{Map-elites Algorithm}{}
\State ($\mathcal{P}\leftarrow \emptyset,\mathcal{X}\leftarrow \emptyset$)
\For {iter $= 1\rightarrow$ I} 
\If {iter $<$ G}
\State $x' \leftarrow $ randomSolution()
\Else
\State $x' \leftarrow $ randomSelection($\mathcal{X}$)
\State $x' \leftarrow $ randomVariation($\mathcal{X}$)
\EndIf
\State $b' \leftarrow $ featureDescriptor(x')
\State $p' \leftarrow $ performance(x')
\If {$\mathcal{P}(b') = \emptyset $ or $ \mathcal{P}(b') < p'$ }
\State $\mathcal{P}(b') \leftarrow p'$
\State $\mathcal{X}(b') \leftarrow x'$
\EndIf
\EndFor
\State {\bf return} feature-performance map($\mathcal{P}$ and $\mathcal{X}$ )
\EndProcedure
\end{algorithmic}
\caption{\label{alg:mapElites} MAP-Elites Algorithm, taken directly from \cite{Mouret-2015} } 
\end{algorithm}

\subsection{The Evolutionary Algorithm}

The EA uses exactly the same representation and operators as the Map-Elites algorithm.  The EA uses a population size of 100, with 40 children being created each generation. Each child is created by cloning from one parent or crossover using two parent. Parents are selected using a tournament of size 2. A mutation-rate of 0.7 is applied to each child. The children are added back into the population, replacing the loser of a tournament, providing the child represents an improvement over the loser. The parameters for the EA were derived from the authors' previous experience with similar algorithms applied to the same problem instances.

\subsection{Problem Instances}
\label{sec:probInstances}
We use a set of problem instances based upon the city of London, divided into two problem sets, termed London (60 visits) and BigLondon (110 visits). 
These instances were first introduced in \cite{Urquhart-2015}. Each visit represents a real post-code within London.
For each of the  problem sets, 5 instances are produced in which the duration of each visit is fixed to 30 minutes. Visits are randomly allocated to  one of  $t$ time-windows, where $t  \in \{1,2,4,8\}$. For $=1$, the time-window has a duration of 8 hours, for $t=2$, the time-windows are ``9am-1pm'' and ``1pm-5pm'' etc.  These instances are labelled using the scheme {\em <set>- numTimeWindows},  i.e. {\em Lon-1} refers to an instance in the London  with one time-window and {\em Blon-2} refers to an instance of the BigLondon problem with 2 time windows.  The fifth instance represents a randomly chosen mixture of time windows based on 1,2,4 and 8 hours.

If a journey is undertaken by car, paths between visits and distance is calculated according to the real road-network using the GraphHopper library\footnote{\url{ https://graphhopper.com/}}. This relies on Open StreetMap data\footnote{\url{https://openstreetmap.org/}}. Car emissions are calculated as 140 g/km based upon values presented in \cite{TFL-2009}.   For journeys by public-transport, data  is read from the Transport for London (TfL) API\footnote{\url{ https://api.tfl.gov.uk/}}  which provides information including times, modes and routes of travel by bus and train. Public transport emissions factors are based upon those published by TfL \cite{TFL-2009}.

 \subsection{Experimental Parameters}
The function evaluation budget is fixed in all experiments. We test two values: 1 million evaluations and 5 million, in each case the first 1000 evaluations are used for initialisation.. Each treatment is repeated 10 times on each instance. The best objective (distance) value is recorded for both treatments in each run. We apply Vargha and Delaney's $\hat{A}$ statistic \cite{vargha2000critique} to assess difference between the algorithms. This is regarded as a robust test when assessing randomised algorithms. The test returns a statistic, $\hat{A}$, that takes values between 0 and 1; a value of 0.5 indicates that the two algorithms are stochastically equivalent, while values closer to 0 or 1 indicate an increasingly large stochastic difference between the algorithms.  One of the most attractive properties of the Vargha-Delaney test is the simple interpretation of the $\hat{A}$ statistic: for results from two algorithms, A and B, then is simply the expected probability that algorithm A produces a superior value to algorithm B. We follow the standard interpretation that a value in the range $0.5\pm 0.06$ indicates a small effect, $0.5 \pm 0.14$ a medium effect and $.5 \pm 0.21$ a large effect. In addition we use two metrics to further analyse Map-Elites that are now de-facto in the literature:
\begin{itemize}
\item {\em Coverage} represents the area of the feature-space covered by a single run of the algorithm, i.e. the number of cells filled. For a single run $x$ of algorithm $y$, $coverage = noOfCellsFilled / C_{Max}$ where $C_{Max}$ is the total number of cells filled by combining all runs of any algorithm  on the problem under consideration.

\item {\em Precision} is also defined as {\em opt-in reliability}: {\em if} a cell is filled in a specific run, then the cell-precision is calculated as the inverse of the performance-value (distance) found in the that cell in that run, divided by the best-value ever obtained for cell in any run of any algorithm (as this is minimisation). Cell-precision is averaged over all {\em filled} cells in an archive to give a single precision value for a run.

\end{itemize}

From the perspective  of a planner, this represents the choice of solutions available to them, while precision indicates whether a cell contains a solution that is likely to of potential use to the planner. The averaged precision for a run indicates the overall quality of the solutions produced.
\begin{table}
\small
\begin{tabular}{|c|c|c|c|c||c|c|c|c|c|c|}
\hline
 & \multicolumn{5}{c||}{London Problems} & \multicolumn{5}{c|}{Big London Problems}\\
  \cline{2-11}
	& Lon-1&	Lon-2	&Lon-4&	Lon-8&	Lon-rnd&	Blon-1&	Blon-2	&Blon-4&	Blon8&	Blon-rnd\\
   \hline
ME(1M) vs EA(1M)	&$\leftrightarrow$&	$\downarrow\downarrow\downarrow$	&\ $\downarrow\downarrow\downarrow$ &	$\downarrow\downarrow\downarrow$ &	$\downarrow\downarrow\downarrow$ &	$\downarrow\downarrow\downarrow$ &	$\downarrow\downarrow\downarrow$ &	$\downarrow\downarrow\downarrow$ &	$\downarrow\downarrow\downarrow$ &	$\downarrow\downarrow\downarrow$ \\
ME(5M) vs EA(5M)	&$\uparrow\uparrow\uparrow$ &	$\uparrow\uparrow\uparrow$ &	$\uparrow\uparrow\uparrow$ &	$\uparrow\uparrow\uparrow$	& $\uparrow\uparrow\uparrow$ &	$\downarrow\downarrow\downarrow$ &	$\downarrow\downarrow$	&$\uparrow$ &	$\uparrow$	&$\downarrow\downarrow\downarrow$\\
\hline
\end{tabular}
\caption{\label{VD} Comparison of Map-Elites (ME) to Evolutionary Algorithm (EA) at $n$ million evaluations. Arrows show Vargha-Delaney A Test Effect Size and Direction}
\end{table}
\section{Results}
\label{sec:resIntro}

The first research question aims to compare the performance of MAP-Elites and EA algorithms under different evaluation budgets to determine whether MAP-Elites might be useful in producing a set of acceptable solutions quickly. Two values are tested : the first is relatively small with 1 million evaluations (as in \cite{Urquhart-2015}); the second increases this to 5 million.

\begin{figure}
\center
 \subfloat[London fitness\label{fig:lon5mfitness}]{
\includegraphics[width=0.45\textwidth]{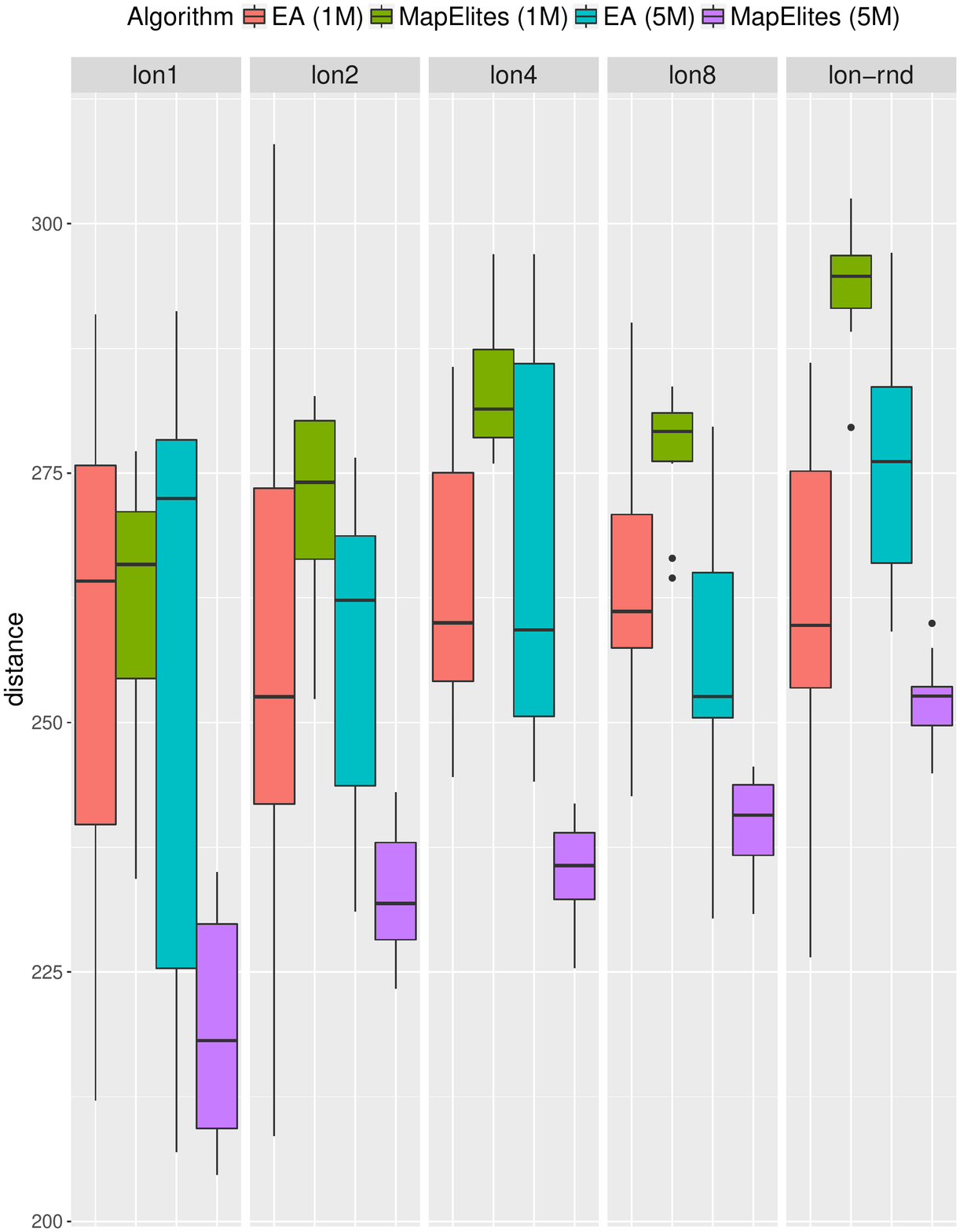}
  }
\subfloat[BigLondon fitness\label{fig:blon5mfitness}]{
        \includegraphics[width=0.45\textwidth]{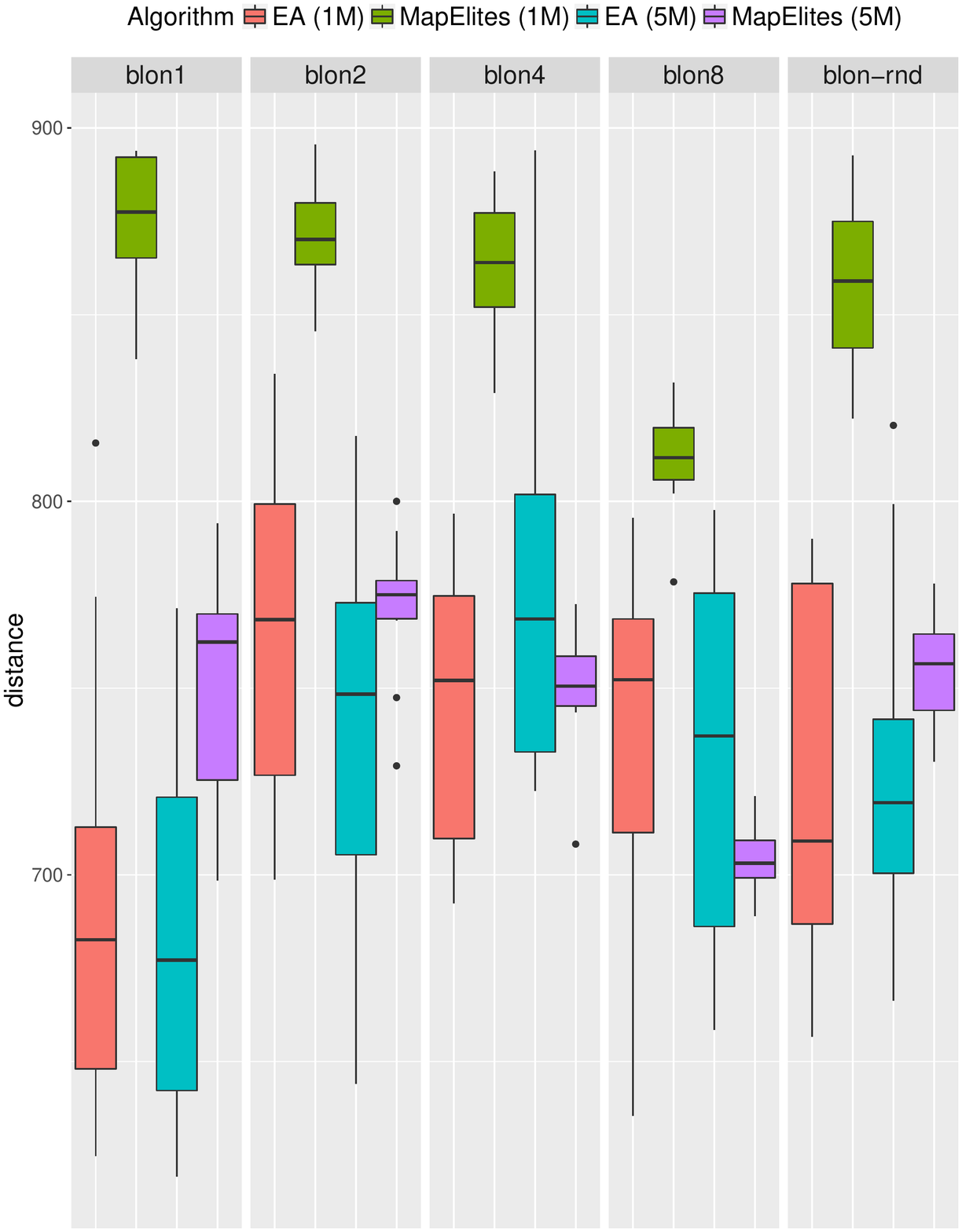}
  }

\caption{\label{budgets} Performance of MAP Elites and the EA with budgets of 1 million and 5 million evaluations.}
\end{figure}

Figures \ref{budgets}(a,b) show the objective fitness values achieved  by ME and the EA under both budgets on each of the problem instances. Table \ref{VD} shows effect size and direction according to the Vargha-Delaney metric.

We note firstly that for 1 million evaluations for both sets of problems, the EA outperforms Map-Elites: the median of the EA is lower than ME, and the effect size is large in each case.  However, when the budget is increased to 5M,  Map-Elites outperforms the EA on all of the smaller problems with a large effect size; it also outperforms the EA on two of the larger problems, although the effect size is small. In the remaining 3 cases, the EA still wins.

Note that the figures \ref{fig:lon5mfitness} and \ref{fig:blon5mfitness} only show performance in terms of distance and do not take into account the four characteristics which provide insight to the end-users. These values are given in table \ref{tab:ME-EA-compare}.  Firstly we note that for the smaller Lon problems, the best-value for each characteristic is obtained from the MAP-Elites algorithm in call cases. This includes Lon-8 in which the best objective value for a solution is obtained by the EA, but the solution has poorer values for each of the 4 characteristics than the best solution obtained by MAP-Elites. Examining the results for the larger BLon problem demonstrates that MAP-Elites, despite a sub-optimal performance (w.r.t the objective function), can still find solutions that out perform the EA in terms of the individual characteristics. We also note that for the Lon problem ME performs equally well regardless of time window constraints. 

\begin{table}
\tiny
\centering
\begin{tabular}{|l|l|l|l|l|l|}

\hline
                  & \textbf{Dist}   & \textbf{StaffCost} & \textbf{TravelCost} & \textbf{CO2}    & \textbf{CarUse} \\ \hline
\textbf{Lon-1}    & {\bf 204.64} : 206.93 & {\bf 841} : 974.67       & {\bf 82.54} : 85.79       & {\bf 133.83} : 163.75 & {\bf 0} : 0.25        \\ \hline
\textbf{Lon-2}    & {\bf 223.3} : 231.02  & {\bf 870.67} : 1014.67   & {\bf 89.71} : 103.04      & {\bf 148.94} : 192.85 & {\bf 0.06} : 0.33     \\ \hline
\textbf{Lon-4}    & {\bf 225.37} : 244.09 & {\bf 904.33} : 1276      & {\bf 94.63} : 116.74      & {\bf 158.77} : 194.59 & {\bf 0.04} : 0.33     \\ \hline
\textbf{Lon-8}    & 230.8 : {\bf 230.34}  & {\bf 967.33} : 1376.67   & {\bf 103.5} : 140.1       & {\bf 159.07} : 240.54 & {\bf 0.04} : 0.35     \\ \hline
\textbf{Lon-Rnd}  & {\bf 244.91} : 259.11 & {\bf 944} : 1140.33      & {\bf 99.48} : 107.4       & {\bf 155.17} : 216.53 & {\bf 0.04} : 0.33     \\ \hline
\textbf{}         &                 &                    &                     &                 &                 \\ \hline
\textbf{Blon-1}   & 698.48 : {\bf 619.15} & {\bf 1987} : 2182.33     & 222.63 : {\bf 207.17}     & 527.27 : {\bf 506.02} & {\bf0.04} : 0.25     \\ \hline
\textbf{Blon-2}   & 729.21 : {\bf 644.07} & {\bf 2107.67} : 2385.67  & 244.54 : {\bf 243.55}     & 584.99 : {\bf 581.16} & {\bf 0.07} : 0.32     \\ \hline
\textbf{Blon-4}   & {\bf 708.25} : 722.53 & {\bf 2183.33} : 2545.67  & {\bf 267.85} : 272.34     & {\bf 584.19} : 637.26 & {\bf 0.08} : 0.33     \\ \hline
\textbf{Blon-8}   & 688.94 : {\bf 658.52} & {\bf 2209} : 2772        & {\bf 272.22} : 311.52     & {\bf 586.81} : 637.5  & {\bf 0.08} : 0.38     \\ \hline
\textbf{Blon-rnd} & 730.3 : {\bf 666.29}  & {\bf 2256} : 2717.67     & {\bf 251.31} : 263.1      & {\bf 580.16} : 602.47 & {\bf 0.09} : 0.36     \\ \hline
\end{tabular}
\caption{Best results found for performance (distance) and the 4 solution characteristics (5 million evaluations). Values are shown for MAP Elites  on the left and the EA on the right.}
\label{tab:ME-EA-compare}
\end{table}

\subsection{Coverage and Precision}
The coverage metric evaluates the ability of an individual run of an algorithm to place individuals in each of the cells. Note that it is possible that some of the cells cannot be filled in because the characteristics of that instance do not allow a feasible solution in that area.

\begin{figure}%
\centering
\subfloat[][]{%
\label{fig:a}%
\includegraphics[width=2in]{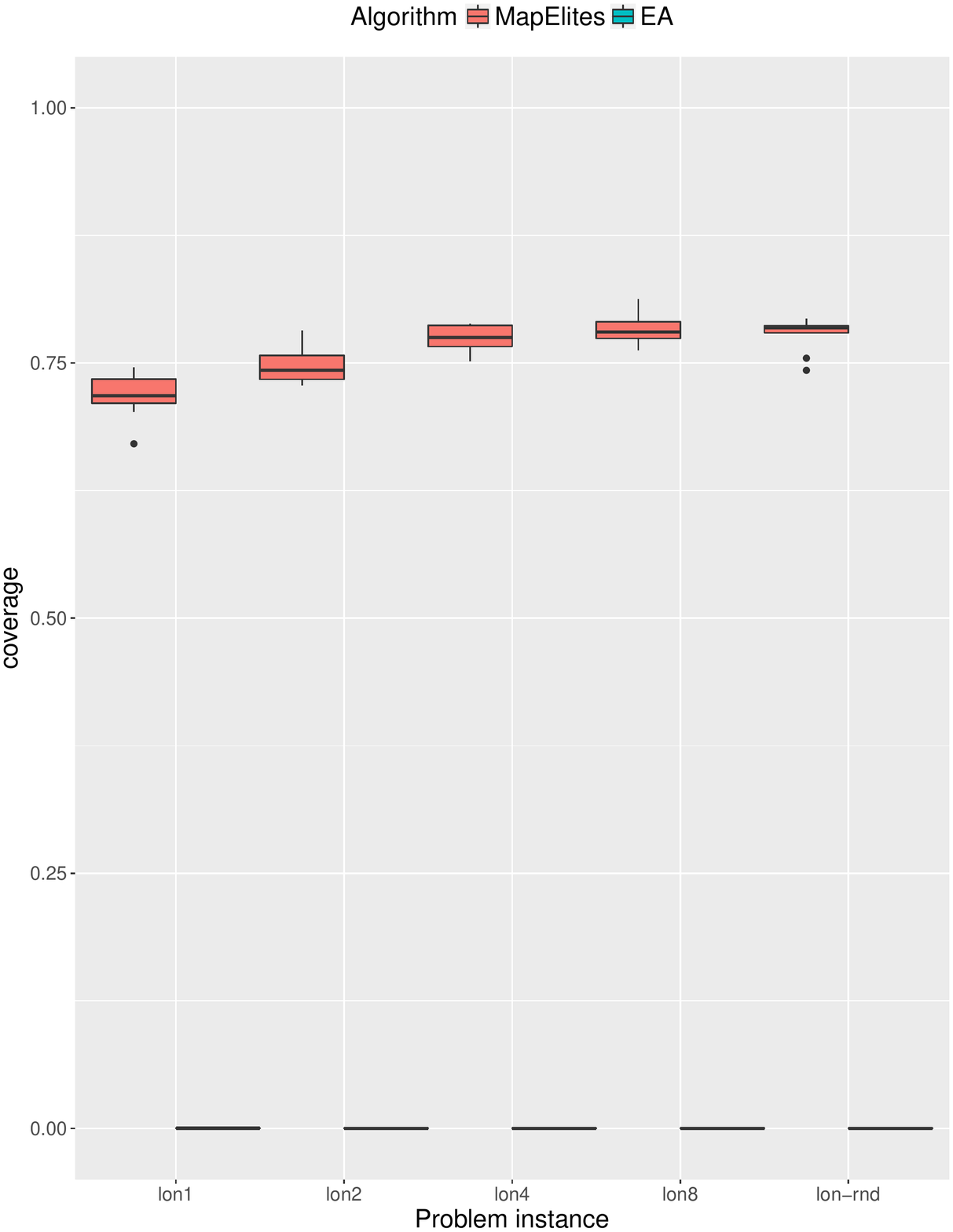}}%
\hspace{8pt}%
\subfloat[][]{%
\label{fig:b}%
\includegraphics[width=2in]{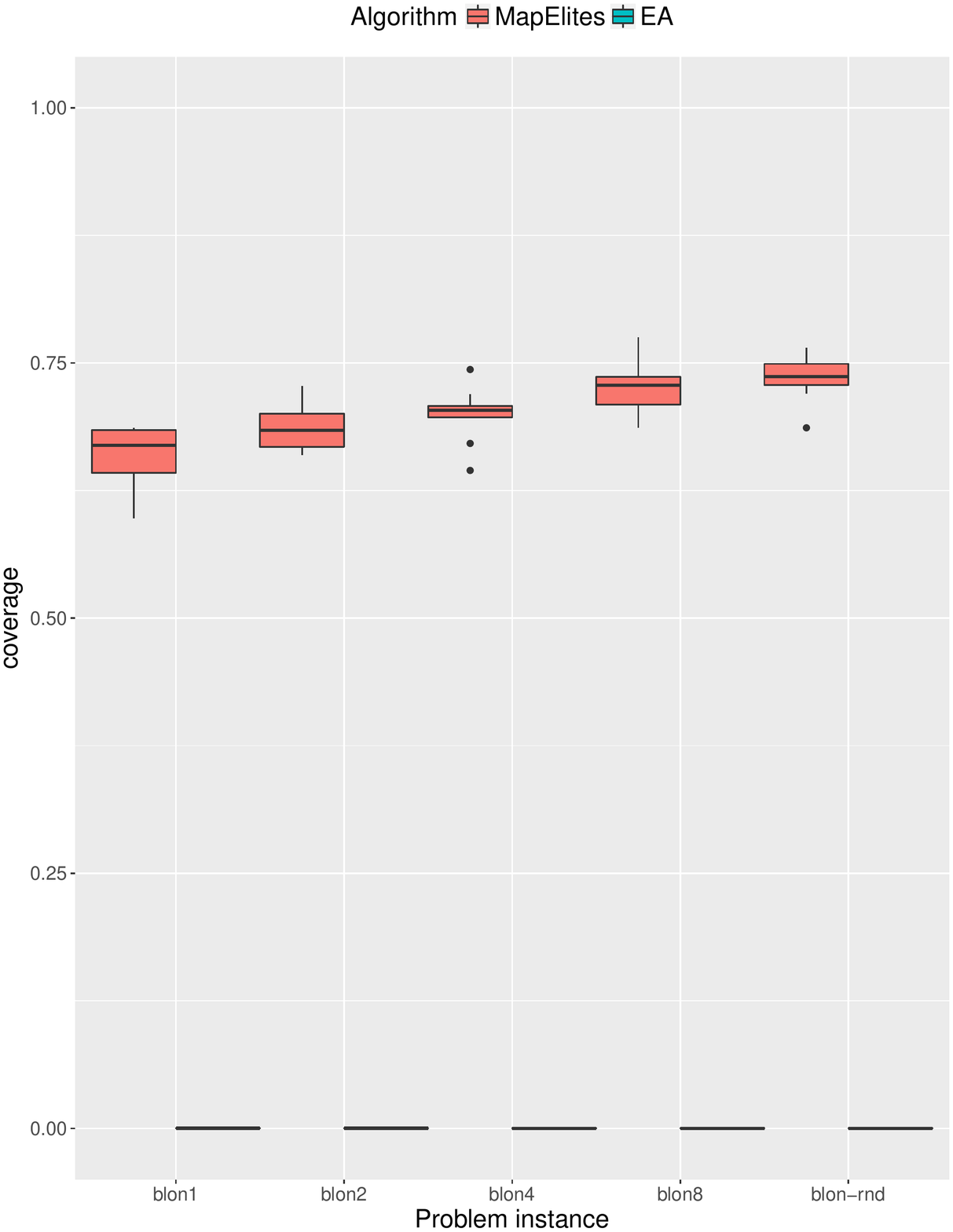}} \\
\subfloat[][]{%
\label{fig:c}%
\includegraphics[height=2in]{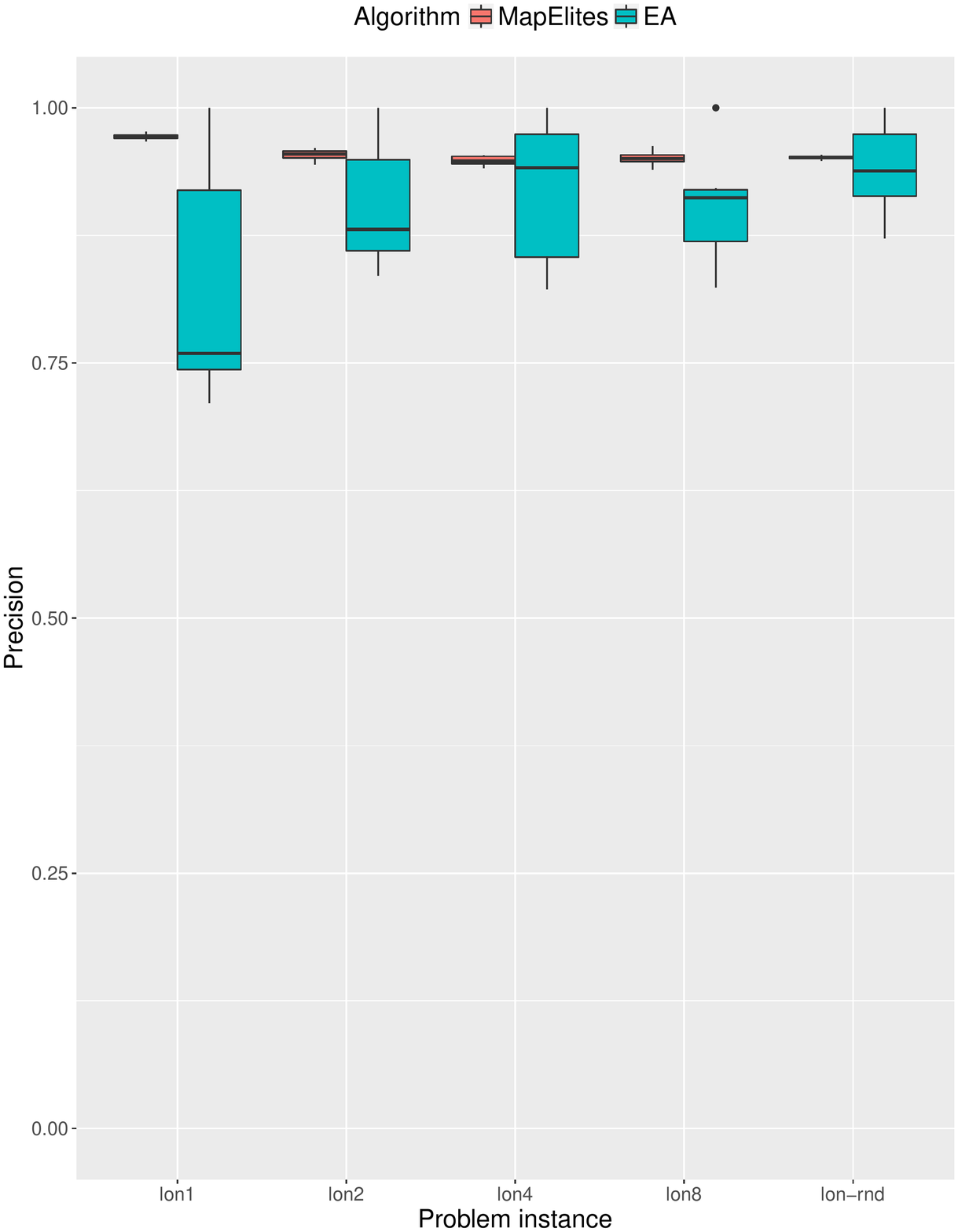}}%
\hspace{8pt}%
\subfloat[][]{%
\label{fig:d}%
\includegraphics[height=2in]{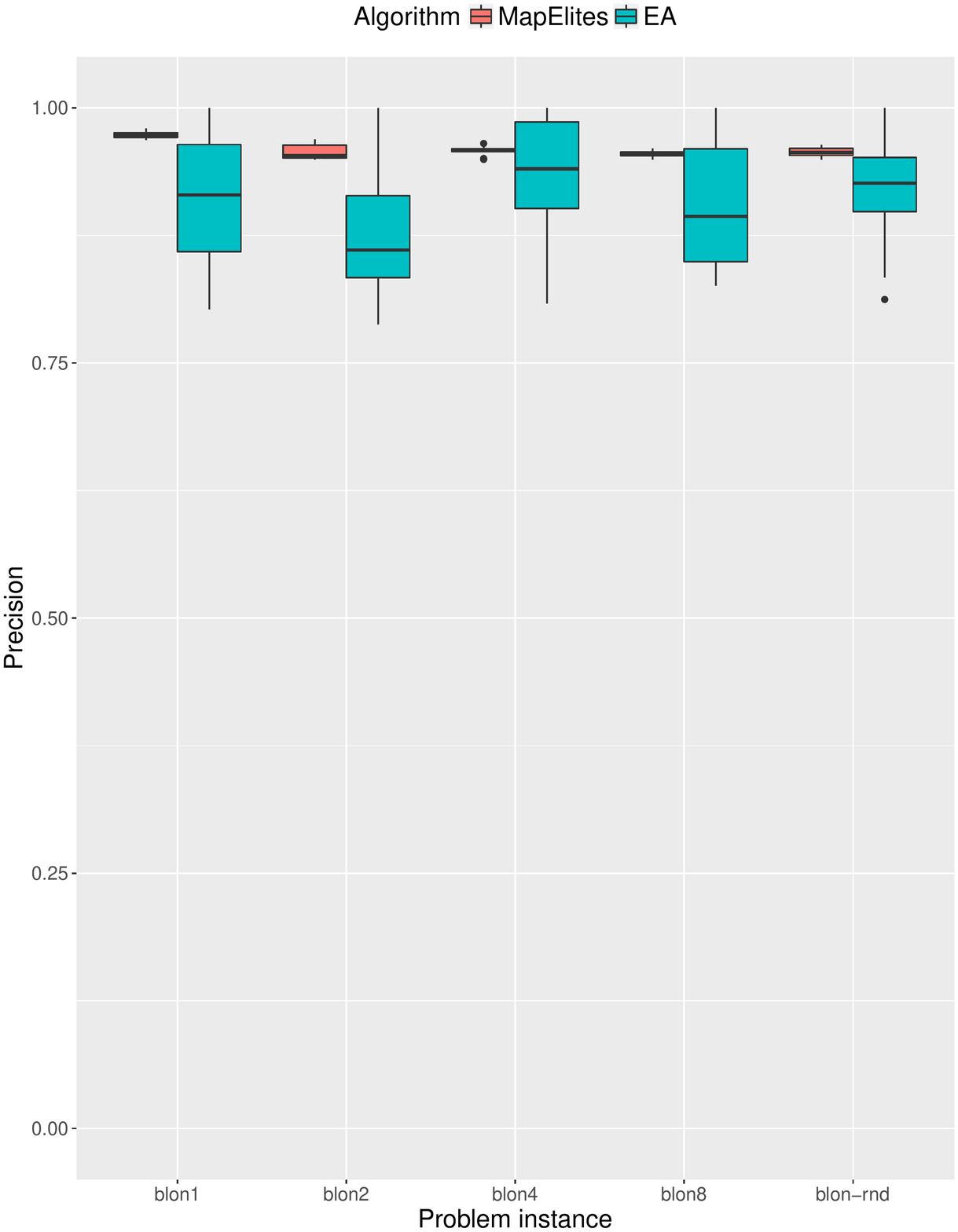}}%
\caption[A set of four subfigures.]{Coverage and Precision for Map-Elites and the EA on both problem sets}%
\label{fig:cov-pre}%
\end{figure}

The coverage achieved is displayed in figures \ref{fig:a} and  \ref{fig:b}. Observe that coverage of over 70\% is common with MAP-Elites, but the EA gives very poor coverage as it converges to a single solution. In real-world terms, the EA leaves the user with little choice of solution and no insight into the problem.

Figures \ref{fig:c} and \ref{fig:d} show the precision achieved by MAP Elites and the EA. Note that the highest precision achieved by the EA outperforms ME, note that precision is calculated over only those cells that are filled.  The EA allocates all of its evaluations to very few cells, and thus find good solutions for those cells. In contrast, MAP-Elites has to distribute the same budget of evaluations across a much larger number cells, making it hard to always find a high-performing solution in each cell. In addition,many of the low-precision scores for ME occur when one run does not find as high-performing a solution in a cell as another run of MAP-Elites.  Running ME for more evaluations would likely improve precision (without danger of convergence due to its propensity to enforce diversity).

\subsection{Gaining Insight into the Problem Domain}

\begin{figure}

\centering
\subfloat[$CO_2$:car use \label{fig:blon-rnd-2d-co2-car}]{
       \includegraphics[width=0.2\textwidth]{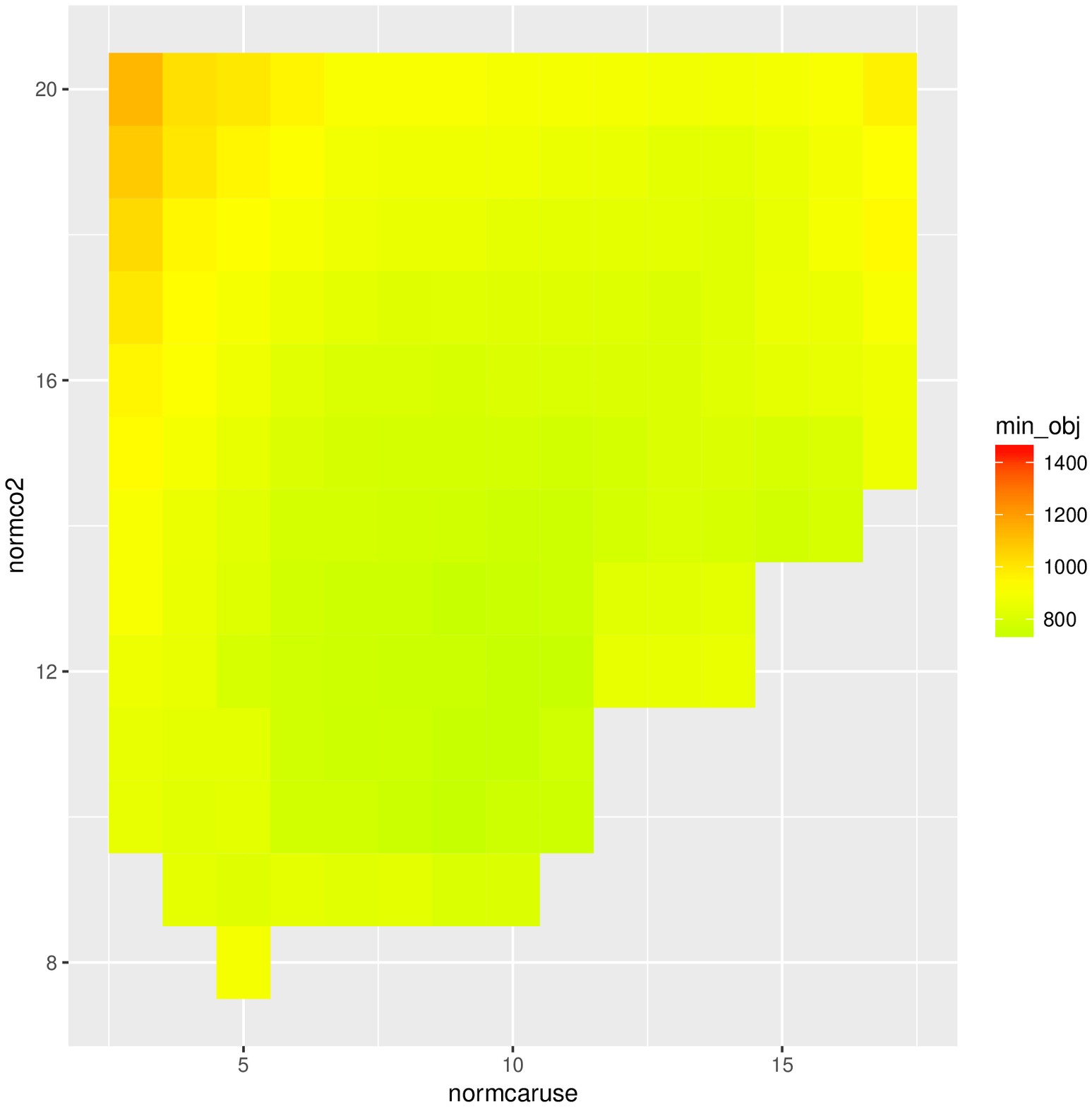}
  }
 \subfloat[$CO_2$:staff cost\label{fig:blon-rnd-2d-co2-staff}]{
       \includegraphics[width=0.2\textwidth]{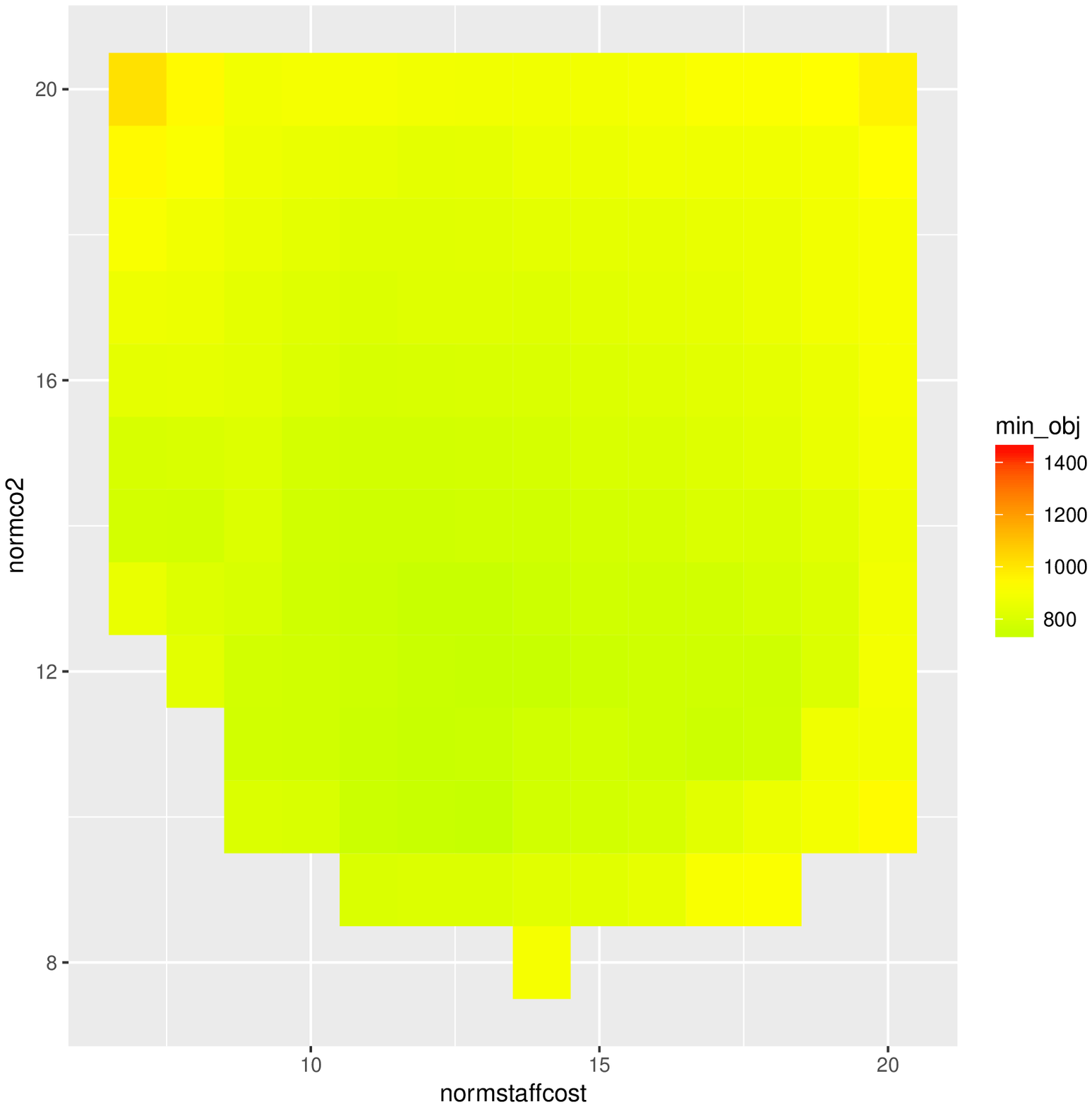}
  }
 \subfloat[staffcost:car use\label{fig:blon-rnd-2d-staff-car}]{
       \includegraphics[width=0.2\textwidth]{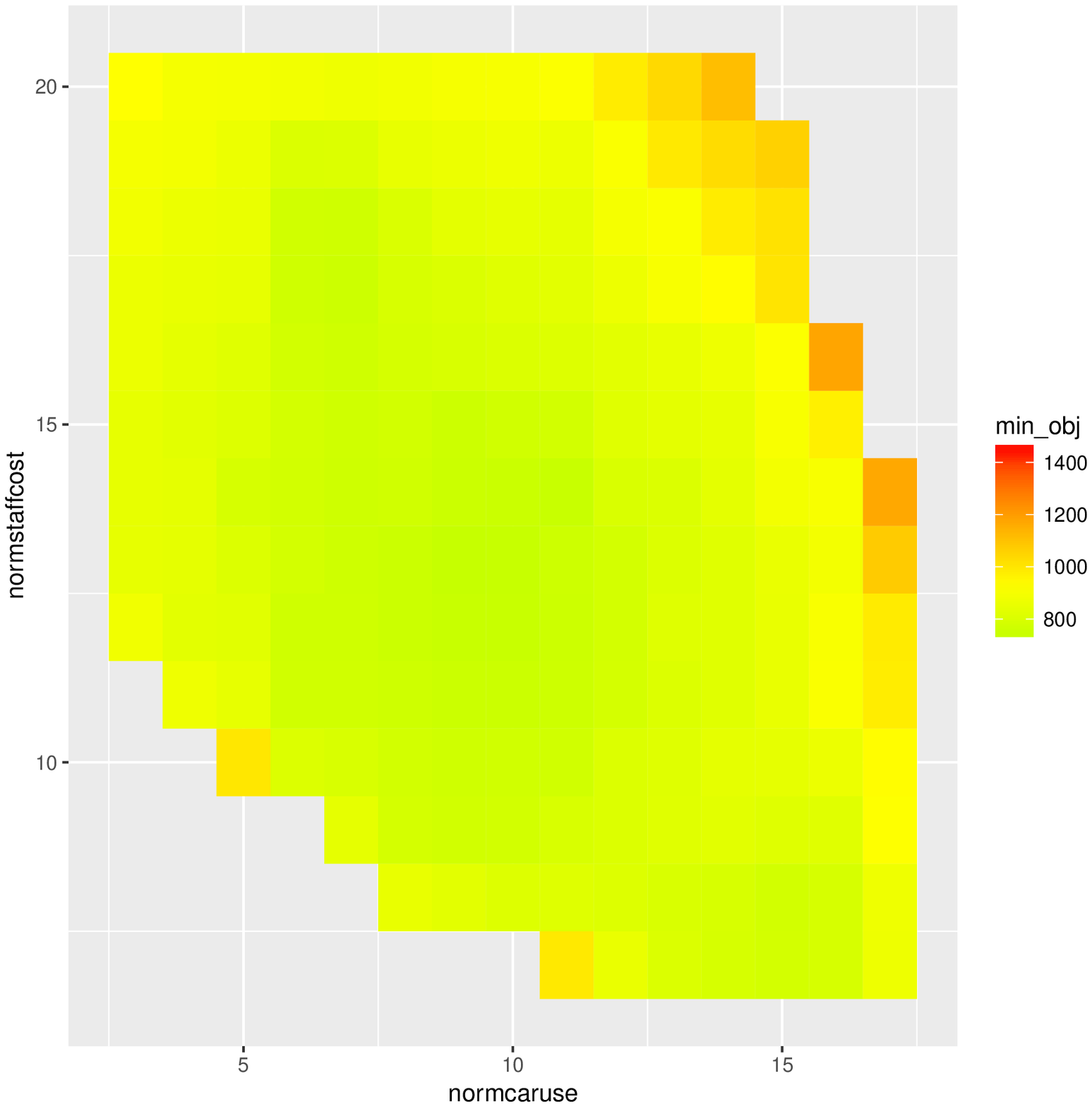}
  }

\subfloat[travelcost:car use\label{fig:blon-rnd-2d-travel-car}]{
       \includegraphics[width=0.2\textwidth]{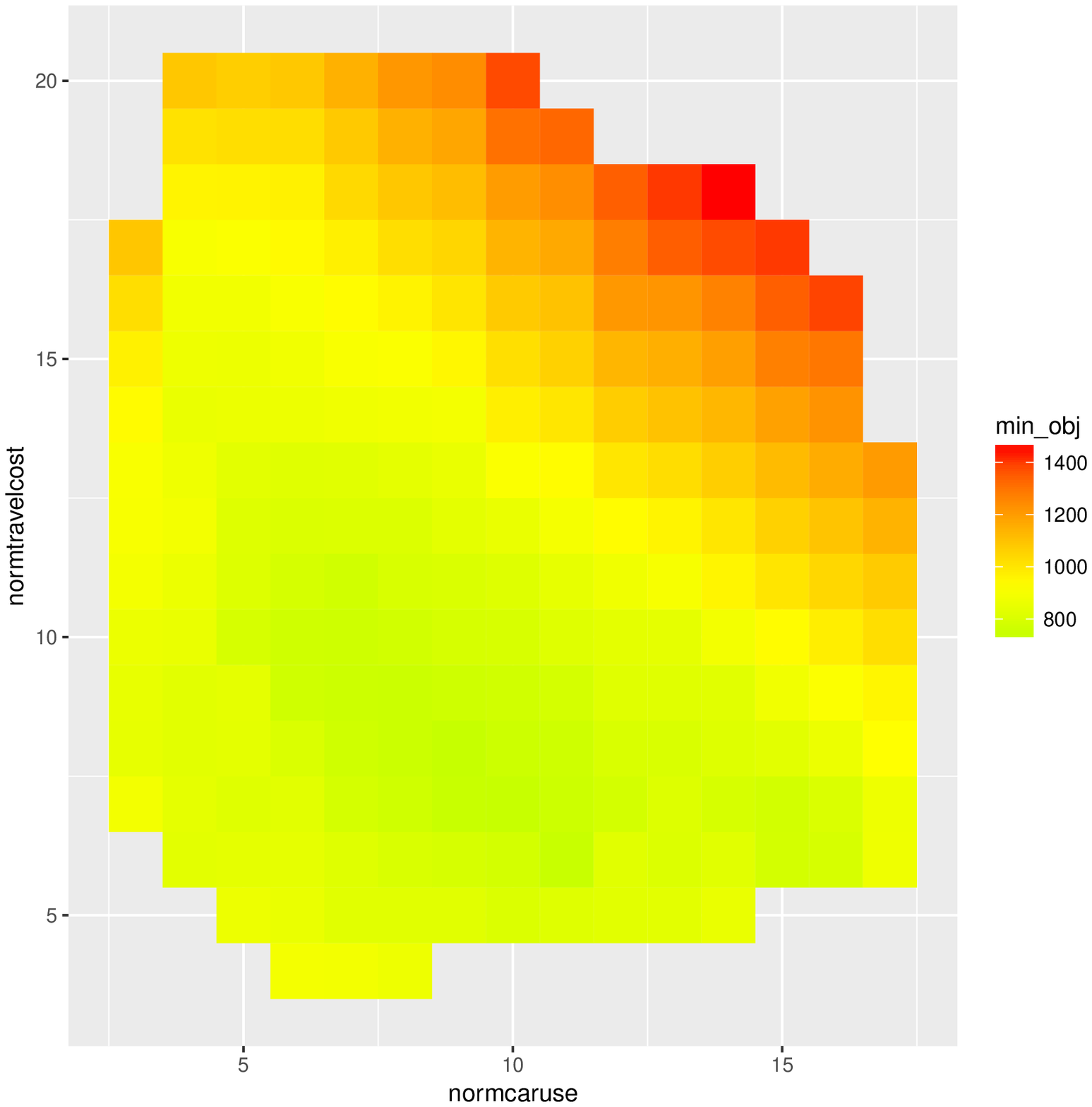}
  }
\subfloat[travel cost:$CO_2$\label{fig:blon-rnd-2d-travel-co2}]{
       \includegraphics[width=0.2\textwidth]{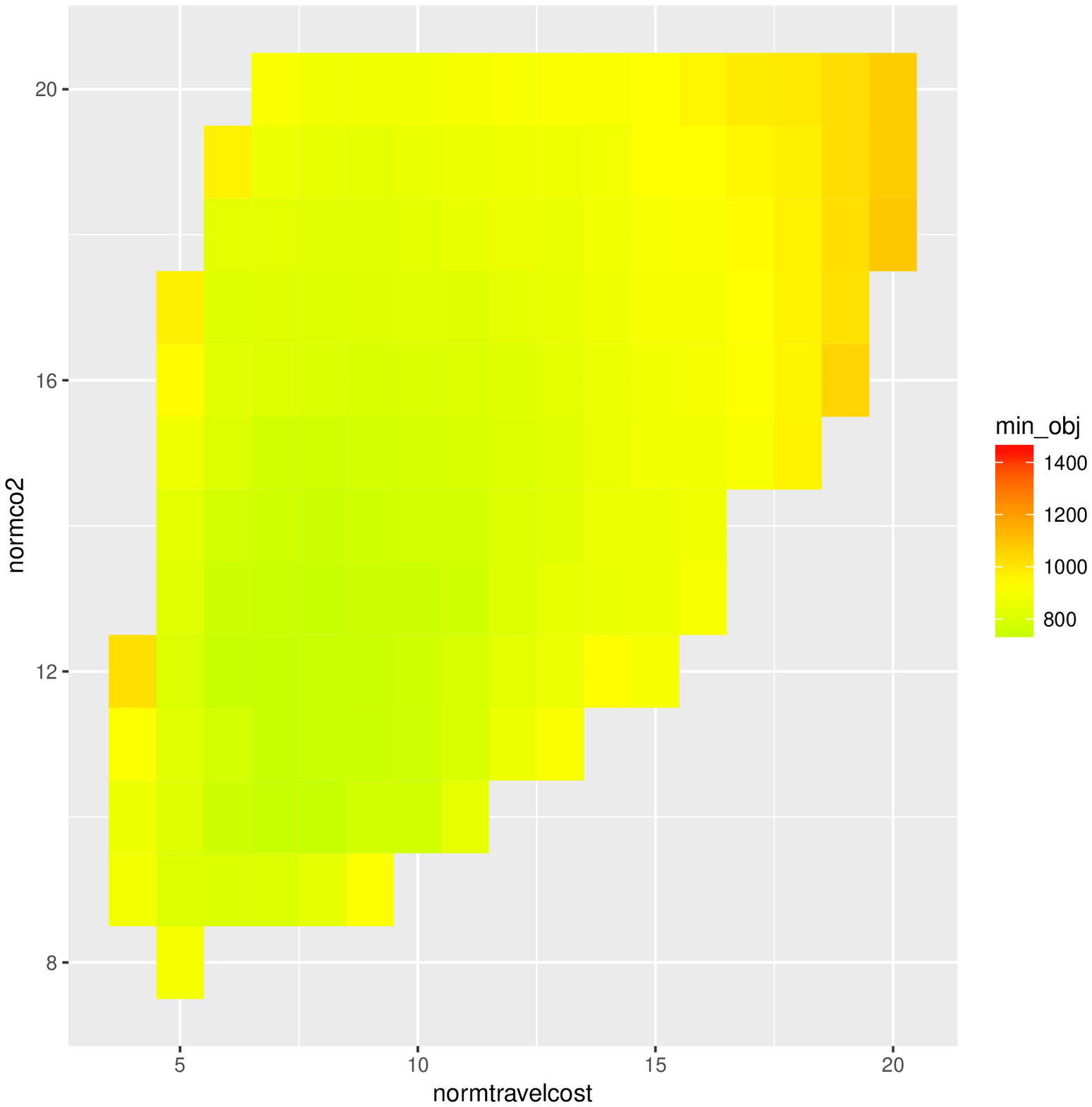}
  }
\subfloat[travel cost:staff cost\label{fig:blon-rnd-2d-travel-staff}]{
       \includegraphics[width=0.2\textwidth]{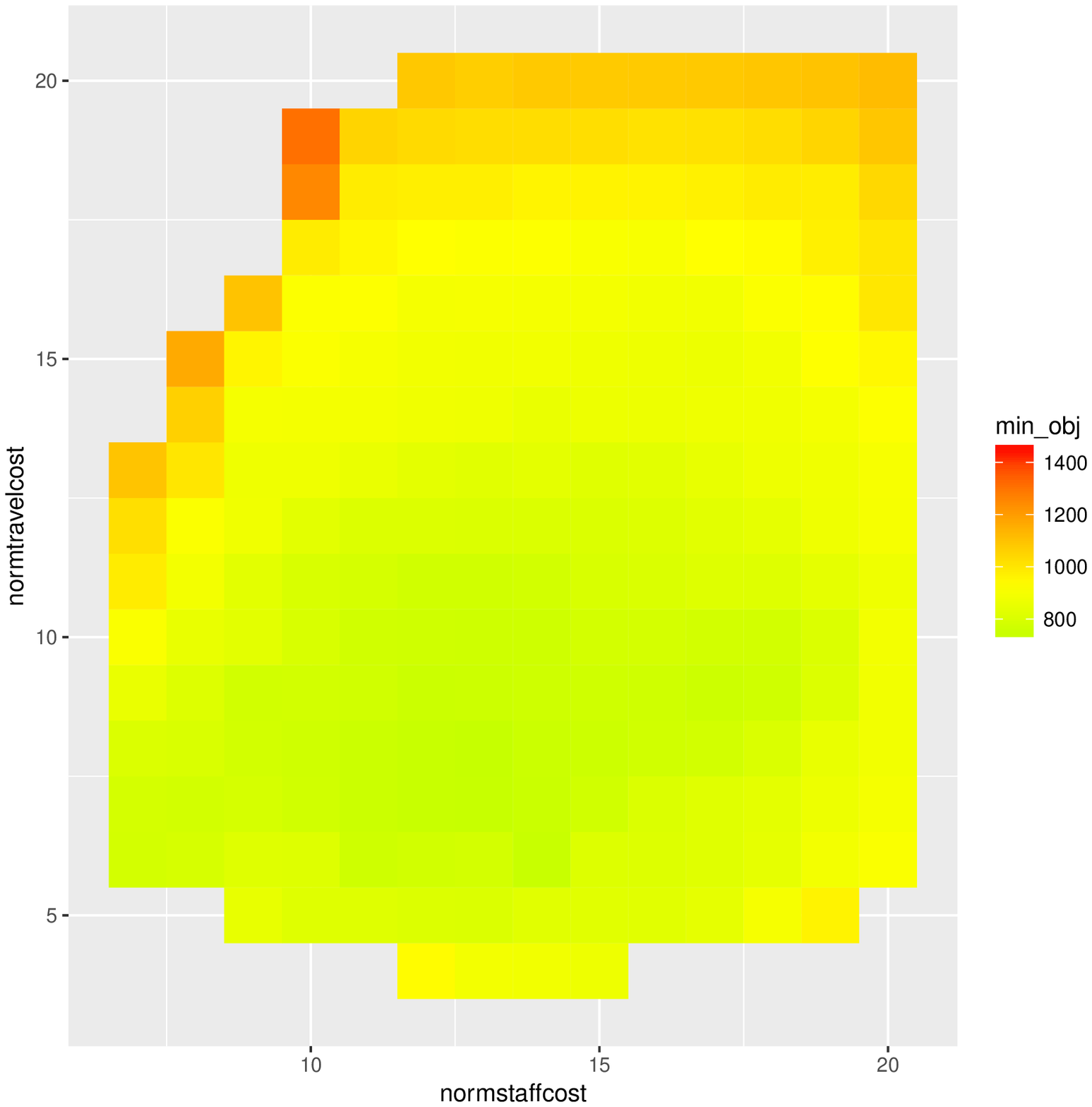}
  }

 \caption{Maps produced from a single run of the Blon-1 problem: (all possible pairings of the 4 characteristics)}
 \label{fig:bloninsight}
 \end{figure}

Figure  \ref{fig:bloninsight} plots the cells, and the elite solutions contained, for each 2-dimensional pairing of the 4 dimensions. Although the archive could be drawn in 4-dimensions, discussion with users suggested that presenting 2-dimensional maps provides more insight. Within each plot, each cell that is occupied is coloured to represent the distance objective value of the elite solution - lowest (best) values being green, highest being red. Note that most of the cells have a solution within them. Where there is an area with no solutions it tends to be at a corner of the plot. For instance, there are a lack of solutions with low $CO_2$ and high travel costs (figure \ref{fig:blon-rnd-2d-travel-co2}) or high car use and low $CO_2$ (figure \ref{fig:blon-rnd-2d-co2-car}). From a planning perspective, figure \ref{fig:bloninsight} indicates (1) combinations of objectives that have no feasible solutions, and (2) quality of feasible solutions.

MAP-Elites tends to cover a larger part of the solution space. A common trend is that the solutions that are better in terms of one or two of the four characteristics are not always solutions that exhibit the lowest distance objective. The map also quantifies trade-offs in objective value: for example,
 the extent to which increased car use increases $CO_2$ compared to options that utilise more public transport. Another insight to be gained is the effects of higher public transport use (i.e. low car use) and staff cost: staff costs rise as public transport usage increases (figures \ref{fig:blon-rnd-2d-staff-car} a). This is due to the longer journey times experienced with  public transport leading to increased working hours for staff.

A planner with responsibility for determining policies regarding staff scheduling may make use of the diagrams in figure \ref{fig:blon-rnd-2d-staff-car} to understand what solutions are possible given a specific priority. For instance, if it is determined that reducing $CO_2$ is a priority then they can determine what possible trade-offs exist for low $CO_2$ solutions. Where a balance is required (i.e. lowering $CO_2$ but also keeping financial costs in check) MAP-Elites allows the planner to find compromise solutions that are not optimal in any single dimension, but may prove useful when meeting multiple organisational targets or aspirations.

\section{Conclusions}
In this paper we have applied MAP-Elites  to a real world combinatorial optimisation problem domain --- a workforce scheduling and routing problem. Unlike previous applications of MAP-Elites that have tended to concentrate on design  problems, WSRP is an example of a repetitive problem, requiring an optimisation algorithm to find acceptable solutions in  a short period of time. In addition to an acceptable solution however, a user also requires choice, in being able to select potential solutions based on additional criteria of relevance to a particular company.

With reference to the research questions in section \ref{sec:intro}, we note that MAP-Elites tends to require a larger evaluation budget to produce results that are comparable with a straightforward EA  for the problems tested. However, for small problems,  affording a larger evaluation budget to Map-Elites enables it to  discover improved solutions, compared to the EA. For larger problems, although our results show that MAP-Elites cannot outperform the EA in terms of objective performance, it does find solutions that outperform the EA in terms of the individual characteristics.  It is likely that running MAP-Elites for longer would continue to improve its performance, without risking convergence. The increased cpu-time required for such a budget may be easily obtained through the use of multi-core desktop computers or cloud based resources in a practical setting.  We also note that the illumination aspect of MAP-Elites may aid the ability of  planners to understand the factors that lead to good solutions and subsequently influence policy planning/determine choices based on organisational values, and that this aspect is of considerable benefit.  Illumination of the solution-space also provides additional insight to planners, who can gain understanding into the influence of different factors on the overall cost of a solution.

Future work will focus on further exploration of the relationship between objective quality and function evaluations, to gain insight into the anytime performance of Map-Elites, for use in a real-world setting. The granularity of the archive clearly influences performance and should be investigated by depth. Finally, an additional comparison to multi-objective approaches is also worth pursing --- while this may improve solution quality however it is  unlikely to offer the same insight into the entire search-space.

\bibliographystyle{splncs04}
\bibliography{refs} 

\end{document}